\documentclass[12pt]{amsart}

\usepackage{amsmath}
\usepackage{amsfonts}
\usepackage{amssymb}
\usepackage{amsthm}

\newtheorem{remark}{Remark}

\newtheorem{question}{Question}

\DeclareMathOperator{\alg}{alg}
\title{Recursion and evolution: Part II}
\author{A. D. Arvanitakis}
\email{aarva@math.ntua.gr}
\address{National Technical University of Athens\\ Department of Mathematics\\ Athens, Greece}

\begin{document}

\begin{abstract} We examine the question of whether it is possible for a diagonalizing system, to learn to use environmental reward and punishment as an information, in order to appropriately adapt.

More specifically, we study the possibility of such a system, to learn to use diagonalization on the basis of a rewarding function.

Relevant phenomena regarding memory are also investigated.
\end{abstract}

\maketitle

\section{Introduction}

A self-editing algorithm is one that can edit its {\em code}. By the term {\em code} we mean a description of the algorithm in the sense of a program for it, together with necessary data. So a self-editing algorithm may alter its program and on the other hand it is determined by it. This interesting duality lies on the core of our study. Notice that the above consideration implicitly assumes an environment that is able to perform computations of some basic set of instructions. (For example, the set of instructions of a programming language). 

Throughout the present study we will assume that the natural environment is able to compute such a basic set of instructions, by means of chemical reactions. Yet another model for self-editing algorithms may as well be in a computer which additionally can provide us with the necessary language to study their behavior.

Assuming such an environment, we denote by $\alg(c)$ the algorithm that corresponds to a code $c.$  We say that the code $c$ is self-editing, if $\alg(c)$ is.

The property of self-editing implies some interesting features about potential capabilities for such algorithms: One can easily see that just by editing their code, they are able to proliferate, (for example by duplicating their code, or by duplicating it somehow altered), or to keep memory of preceding codes. 

Proliferation imposes to the subject of our study, the structure of a tree. If the sequence
\begin{equation} \label{sequence} c_1 \mapsto c_2 \mapsto \cdots \mapsto c_n \end{equation}
is a branch of such a tree, a code $r$ is said to {\em fit} the sequence in \eqref{sequence}, if 
\[ \alg(r)(c_i) = c_{i+1}, \quad i=1, 2, \dots, n.\]

As mentioned earlier, self-editing algorithms may retain the memory of their previous codes, so that we may assume that the actual input of $\alg(c_n),$ is the all sequence in \eqref{sequence}. Under this consideration, we may define a {\em diagonalization} instruction to be the procedure during which $\alg(c_n)$ finds a code $r$ that fits the sequence in \eqref{sequence} and uses it thereafter to compute its descendants. 

At this point, diagonalization seems to have not any meaning at all, except to reproduce any simple logical pattern that is by chance existed in \eqref{sequence}. However, assuming some kind of natural or artificial selection, \eqref{sequence} is automatically rendered as a {\em surviving} sequence and thus any such simple logical pattern may reveal this way an environmental demand in which \eqref{sequence} is forced by selection to adapt. Diagonalization in this case may be thought of as a form of perpetuating this adaption. A lot of unexpected results can follow this simple principle, especially when combined with self-editing. Some of them we have already demonstrate in \cite{partI}.

The present paper, deals mainly with the following question:

\begin{question} \label{basicQuestion} Is it possible for a diagonalizing system to evolve so that to establish diagonalization on the basis of a rewarding function? \end{question}

Let us explain a little better question \ref{basicQuestion}. As we saw earlier, the success of diagonalization depends heavily on the relative success of the codes that participate in the sequence to be diagonalized. (The codes $c_i,$ $i= 1, 2, \dots, n$ of a surviving sequence like \eqref{sequence}.) Selection plays a very important role here, exactly by choosing these successful codes. Yet on a more delicate system, decisions should be made not merely on the basis of what is surviving, but also on the basis on what is rewarded. Furthermore punishment brings into consideration the potentiality of drawing information not only from what goes right, but also from what goes wrong. So question \ref{basicQuestion} can be rephrased as: Can a diagonalizing system learn to draw information from reward and punishment? 

Section \ref{section2} is a short introduction to {\em stepping back} instructions and their use, that is possible deletions or deactivations of one or more recursors. Notice that survivability of a sequence implies that such instructions have actually been activated whenever a false recursor has been used.

It is interesting that stepping back transitions may be also thought of as resulting from a non-symmetric proliferating transition, as explained in more detail in paragraph \ref{non-symmetric}.

Stepping back transitions is a subject that is intended to be investigated further later on.

Section \ref{section3} is devoted to the investigation of our main question \ref{basicQuestion}.

We examine various implications of a rewarding function:

{\em Punishment reaction} is a spontaneous deletion or deactivation of a newly employed recursor, under a punishment.

{\em Fading memory} is the act of neglecting codes from the actual history during the procedure of forming a sequence to be diagonalized.

Finally, {\em negative diagonalization} may be thought of as a means to deduce information about rejecting candidate recursors, out of a sample of already rejected ones.

Other than \cite{partI}, the paper has no other prerequisites.  However the ideas involved are closely related to {\em recursion,} which goes back to the ancient Greeks, {\em self-reference} and the {\em diagonal method} which have been introduced by Cantor and studied by various mathematicians such as Church, G\"odel, Kleene, Tarski, Turing etc. The interested reader may find relative material in any context of Set theory, Logic and Recursion Theory such as \cite{moschovakis, kleene}.

Finally, it is well known that the idea of evolving by means of proliferating combined with selection is due to Darwin.

During the research presented in this paper, many people contributed to it, in various ways. (In chronological order) Maria Avouri \cite{maria}, James Stein \cite{jim}, Despoina Zisimopoulou \cite{despoina}, Dimitris Apatsidis \cite{mitsos}, Fotis Mavridis \cite{fotis}, Antonis Charalambopoulos \cite{antonis1}, Vanda Douka \cite{vanda}, Antonis Karamolegos \cite{antonis2}, Helena Papanikolaou \cite{helena} and Miltos Karamanlis \cite{miltos} are among them.

\section{A short introduction to Cycles} \label{section2}

\subsection{Recursors as modifiers} Assume  
\begin{equation} \label{1new} c \mapsto s \end{equation}
to be a self-editing transition. (I.e. $s = \alg(c)(c)$). It is easy to see that in \eqref{1new}, for any codes $c$ and $s,$ we may assume that $s$ is of the form $(r,c),$ where $r$ is an appropriately chosen modification of the code $c.$ 
 A somewhat clumsy but very general way to see this, is by defining $r = r(s)$ to be a code for the algorithm:
\[ \textsf{Use $s$ instead of $c.$} \]

Although this is not demonstrated by the above  construction of the code $r,$ generally, having in mind that the basic code $c,$ is a sufficiently evolved code,  $r$ could be thought of as a {\em minor} change to the main code $c,$ so that representing a transition with an application of the appropriate modifying code ($r$ in this case) to alter the older one, has the main advantage of a more compact memory sequence, provided that in the structure $(r,c),$ one is able to distinguish $r$ as being the newest sub-code. In what follows, we are going to assume such a rating of modifying codes from older to newer ones.

Notice that the modifying code $r$ may depends upon (having as input) $c,$ (the previous state), or not. Therefore it may not induce recursion in the strict meaning of the term, instead it may simply alter the main code $c.$ Yet, for purposes of convenience in terminology, we will insist calling it a {\em recursor} in any case. Thus in our terminology, a recursor might as well be just a modification of a given code (depending or not on the previous states of the code).

Other than the one of more compact memory, there is a dual advantage in such a representation:  In the structure $(r, c),$ $r,$ (the newer modifying code), can be thought of as an {\em under testing} modification of $c.$ Although {\em newest} modifications are not necessarily the ones to blame if something suddenly goes wrong, this is certainly a rule that all the same has some exceptions. Thus, representing a transition rather as an application of an appropriate recursor, includes the advantage of a {\em primitive form} to rate parts of a given code. 

Let us recall that we have already seen a simple version of this general rule, by applying diagonalization to decide whether to copy a particular part of the code
or not. (This was the 1st example in paragraph 3.1.1 of \cite{partI}). The success of this example depends heavily on the hypothesis that a persisting and unaltered for a long time code, should be the correct to use thereafter.

\subsection{Stepping back transitions} In what follows, we will assume the use, as basic instructions, of {\em deletion} or {\em deactivation} of a sub-code. For purposes of a unified approach we will deal with both simultaneously by referring to them as {\em stepping back transitions.} 

It is evident that the algorithm of a self-editing code may  delete or deactivate any part of the code, performing thus a stepping back transition. Such a transition may be performed either as a part of a more general computation, or assuming a non-deterministic nature of our system, simply by chance. 

\subsubsection{Proliferating stepping back transitions} 
\label{non-symmetric}
Interestingly enough, a stepping back transition may be a transition that affects only a part of descendants. For example consider the following simple non-symmetric proliferating computation:
\begin{equation} \label{nonSymmetricProliferation} (r, c) \mapsto \{ (r, c_1), c \}. \end{equation}
In the above transition, we assume that $r$ is a recursor and the code $(r, c)$ gives two descendants, namely $(r, c_1)$ and $c,$ for which $c_1 = \alg(r)(c).$ Employing the technique described in paragraph 2.3 of \cite{partI}, about self-editing computations, it is easy to see that if $(r, c)$ is a self-editing code, then \eqref{nonSymmetricProliferation} can be triggered by the algorithm of $(r, c)$ itself.

We will call proliferating transitions like \eqref{nonSymmetricProliferation}, as {\em non-symmetric.} 

Since in transition \ref{nonSymmetricProliferation}, the denoted by $(r,c_1)$ descendant continues using the recursor $r,$ while the one with code $c,$ not, we may deduce that non-symmetric proliferation can be of great value for using one or more recursors cautiously. 

Its contribution to learning via stepping back transitions is also of great importance, yet we plan to examine it later on.

\subsubsection{} Assuming an environment which is close to the natural one (and therefore to the one we know), the cases of generally valid rules governing it, is rather an exception. On the opposite, local rules are very often met. Therefore it should be the case that in a surviving sequence 
 \begin{equation} \label{1} c_1 \mapsto c_2 \mapsto \cdots \mapsto c_n, \end{equation} there should be transitions of the form of applying (probably by diagonalization), a new recursor $r$, that are followed after a while by stepping back transitions that either delete or  deactivate it. 

We will call the sequence interval between two successive stepping back transitions, an {\em attempt} or a {\em cycle} of the system.

\subsubsection{} Examples showing the necessity of stepping back transitions, may be constructed by using experiments in a sequence, with no sequential correlation between them. If for example a recursor $r_1$ is needed to solve experiment $a_1,$  and $a_2$ is the successive to $a_1$ experiment, then it might be the case, (as it was usually in experiments of \cite{partI}), that a modification of $r_1$ would solve $a_2,$ but it might also be the case that $a_1$ and $a_2$ are of completely different logic, so as in order to solve $a_2,$ one has to completely discard $r_1$ by deleting or simply deactivating it.

So, a conclusion of the previous argumentation is that 
\begin{remark} \label{2} A surviving sequence as \eqref{1}, should contain stepping back transitions at various points of it. \end{remark}

Let us also observe here, that the same arguments show that by aiming to design such a system in an environment of a computer, one has to take into account the necessity of including {\em deletion} and {\em deactivation} to the basic elementary functions of the system.

\section{Response to reward-punishment} \label{section3}
\subsection{Rewarding functions} It is interesting at this point to introduce to our study a {\em rewarding function}. The term refers to a point system which we use to evaluate the behavior of our system. In natural environments there are usually more than one such point systems, for example one could measure the amount of gathered energy or of any of the essential elements for propagation. In human societies, it is usually represented by financial reward, in education by grades and so on. In a computer environment, such points may  represent for example, rights either to use disk or memory space, or to use the processor of the system.

For the beginning, we are going to assume two maybe over-simplified principles:
\begin{enumerate} \item The accumulated points should be significantly correlated with {\em life} or {\em death} of our codes and
\item The corresponding algorithms should be aware of their rating, that is of the amount of points (being positive or negative) delivered by the rewarding function. \end{enumerate}
The second principle can be accomplished in two ways: 

We either insist that the outcome of the rewarding function is {\em noticeable}, in the sense of the definition contained in paragraph 5.1 of \cite{partI}, (i.e. that the outcome of the function is registered in the code and is accessible via an address), or that the function itself has a simple enough code, so that it is expectable by means of the procedure analyzed in paragraph 5.3 of \cite{partI}, to become noticeable. So, in both cases we may assume that the outcome of the rewarding function is noticeable.

A negative assignment of the rewarding function will be called a {\em punishment}.

There is a very natural and important question that arises here: 

\begin{question} \label{4} Is a diagonalizing system capable of understanding and appropriately utilizing the outcome of the rewarding function? \end{question}

We will try below to examine various cases that are related to this question.

\subsection{Punishment reaction} \label{punishmentReaction} As a first and simple attempt to try to investigate such a question, we may indicate that in a surviving sequence such as \eqref{1}, it should be the case that in a lot of transitions, since the sequence is surviving, stepping back instructions should be the correct decision, so that in many cases, they should follow a negative assignment of the rewarding function. Thus by Remark \ref{2} and by diagonalizing over the sequence \eqref{1}, we conclude that a code for 
\begin{equation} \label{3} \textsf{In case of a punishment, step back}, \end{equation}
should fit the sequence. Concluding, if this same code is simple, the system should be expected to use it thereafter. We should notice here that \eqref{3} should refer mainly and as a general case, to the {\em newest} created recursors, although as we have seen, this is not to be always the case.

A more delicate argument in favor of the establishment of the punishment reaction, is that a poorly rewarded code may have difficulty to maintain  all of its parts, so it is more likely to step back, either in the case of a proliferating transition, in which a non-symmetric one would follow, or in a non-proliferating transition.

All the same, in either way, \eqref{3} should fit the memory sequence, establishing thus the punishment reaction.

Examples of the use of the punishment reaction are rather straightforward: A new and unknown change in the state of a code (represented by a recursor), should be immediately rejected after a negative assignment of the rewarding function. This should be thought of something as a usually correct reflex, since the aforementioned negative assignment may be due to completely different reasons. 

On the other hand, as we are going to see in Remark \ref{searchingBetterTarget}, a punishment reaction is neither always useful, nor it can be guaranteed, especially in cases where its functionality would harm the searching capabilities of our system. 

\subsubsection{} Sooner or later, one is forced by the  need to consider a non-deterministic system. This last involves not only the actions that such a system is likely to take, but also the possible inputs of algorithms run by it. To do this, one has to rate both possible actions as well as possible inputs. This may be done in at least two ways: 

The first, which is very straightforward, but it lacks efficiency, is to consider activated repetitions of various codes. The corresponding probability assignment as a result of doing so, may be calculated on the basis of relative frequencies of the activated repetitions. I.e., for example, if the system has to decide among the codes $s_1$ and $s_2$ to use as inputs to a particular procedure, then activated repetition of $s_1$ by $n_1$ times and activated repetition of $s_2$ by $n_2$ times should imply a probability assignment of $n_1/(n_1 + n_2)$ for the first and correspondingly $n_2/(n_1 + n_2)$ for the second. It is evident that the probability calculated thus, should have to do also with the procedure at hand, something that justifies the use of {\em activated} (and not mere) repetitions. 

The second way considered here is much more efficient and has to do with assigning weights, (as a more advanced way of simulating repetitions), both to possible inputs of a procedure as well as to the possible actions, under given circumstances. This method reminds synapses of the nervous system and has been used with success in machine learning.

We have to observe here that the particular method of representing probability assignments by relative frequencies has the advantage of a straightforward manipulation of the corresponding probability. For example in the case of weights, our system is able to raise the probability of taking an action, by simply increasing the corresponding weight.

One should recall here, that by the methods we analyzed in \cite{partI}, our system can use selection in combination with diagonalization, to understand whether it is appropriate to increase or decrease a weight.

The exposition we used here, about how to implement in practice a non-deterministic system, should be thought of rather as a collection of thoughts to keep us going and certainly not as a full analysis of the subject.

\subsection{Memory fading} We will try here to demonstrate another interesting example of the possible adaptation of a diagonalizing system to the rewarding function, which consists of its {\em fading} or {\em highlighting} the memory which is feeding the diagonalizing procedure.

A rewarding function assumes the potentiality of mistaking without death as an input. Thus it leaves the possibility of relatively failing codes in a surviving sequence, or of relatively failing attempts. This should be apparent by the negative value of the rewarding function. 

On the other hand, the success of diagonalizing over a sequence $(c_i)_{i=1}^n,$ depends heavily on the individual success of each code $c_i$ that participates in it. Up to now we needed not worry about that, since merely to live is a success. Yet for more advanced (and thus more complicated) codes, proliferation is an issue regarding the needed resources to be used for carrying it out. And certainly it would be a great waste assigning death to an advanced code for failing to answer a question that simply does not have the necessary information to deduce its answer. (For example failing to answer correctly during the first steps of experiments in 2nd and 3rd example of \cite{partI}). 

In order to deal with the situation, assume that each $c_i$ participates with a weight $w_i$ in the sequence
\[ c_1 \mapsto c_2 \mapsto \cdots \mapsto c_n \]
 to be diagonalized. Now we may also assume that the values $w_i$ are a subject to (non-deterministic) change in the codes $c_j,$ $j\ge i.$ That is the value of each $w_i$ is in fact a function also of $j,$ $w_i(j)$ for $j \ge i.$ 

Now, by selection, one can judge the success of including $c_i$ in the sequence to be diagonalized, by looking at the values $w_i(j),$ $j=i, i+1, \dots:$ 

If a code $c_i$ successfully participates in diagonalization, then an increased corresponding weight $w_i$ should be more fit, therefore in a surviving branch, the values $w_i(j),$ $j=i, i+1, \dots$ should form an increasing sequence. 

Similarly, if $c_i$ does not successfully participates in diagonalization,  they should form a decreasing sequence. 

Therefore by noticing this sequence one can come to a conclusion whether to include the corresponding code in diagonalization or not. 

Since our system may diagonalize over the sequence $w_i(j),$ $j=i, i+1, \dots,$ it will come to the same judgment, just by filling up the sequence. 

Assuming moreover that the rewarding function has actually punished $c_i$ in case of failing, using a more general diagonal now, it may be seen that a code for
\begin{equation} \label{fadingMemory} \textsf{\begin{tabular}{p{10cm}}In case of a negative assignment of the rewarding function on some code $c_i,$ restrain from using $c_i$ in the sequence to be diagonalized,\end{tabular}} \end{equation}
fits the surviving sequence, so that if simple, it will be followed hereafter.

One can by analogy see that relative failure and fading memory may be respectively replaced by relative success and highlighting memory.

\begin{remark} {\em It is evident that the same conclusion could be carried out without the need of weights and by simple letting our system to decide based on chance, whether or not to include any $c_i$ in the sequence to be diagonalized. We preferred here the analysis with weights, in order to demonstrate the {\em fading memory} effect described thus. 

One should notice also that this effect is irrelevant with the rewarding function itself and has to do only with the resulting success of including a particular code in the sequence to be diagonalized. Applying this observation to sub-codes, as we are going to see later in more detail, can result in deciding of our system to use the most appropriate information in order to diagonalize, in accordance with the task at hand. 

For example, given an address $\theta$ of the code $c,$ in order to decide about editing $c\restriction \theta,$ one usually needs to know only about previous values in $\theta.$ Knowing such a rule, (with exceptions whatsoever), saves time and space and could be accomplished by memory fading as we described it above. (Simply put, irrelevant sub-codes will be subject to memory fading).

Thus 
\[ \textsf{\begin{tabular}{p{10cm}} Use only the past values in $\theta$ to decide about $\theta$-contents, \end{tabular}} \]}
\end{remark}
should fit the sequence.

\subsubsection{Revisiting the examples in \cite{partI}} Experiments in \cite{partI} may be seen now with a fresh eye, on the basis of memory fading caused by punishment (or reduced reward). 

Indeed, let us consider the general case, as an experiment of the form
\[ (a_1, a_2, a_3, \dots)\]
where $a_i,$ $i=1, 2, 3,$ are exactly the non-intended to be guessed codes. 

Up to now we have assumed that in any case of the sub-experiments $(1),$ $(2)$ and $(3),$ in which our system has to guess $a_1,$ $a_2$ and $a_3$ respectively, the guessing takes place by means of an adequate set of descendants, so that at least one of them outputs the correct code. 

It can be easily seen now, that we may replace this assumption by attempts made by our system, that is, without the use of proliferation. 

Indeed, assume that during the attempt of our system to guess sub-experiment $(i),$ it does not produce the correct code and moreover that it is punished for this by the rewarding function. Such attempts, on the basis of \eqref{fadingMemory}, are not going to participate in the sequence to be diagonalized, producing thus the same result of guessing the intended to be guessed codes $a_i,$ $i>3,$ by the diagonalizing procedure, since this procedure will be applied to the (correct) codes that output the (correct) answers $a_i,$ $i=1, 2, 3.$ 

It is evident that one has, as previously done, to rely on the assumption that the codes $a_i,$ $i=1,2,3,$ will be eventually produced. This however complies with the fact that these are not-intended to be guessed codes.

\begin{remark} \label{searchingBetterTarget} {\em Although we tried here to demonstrate how diagonalization can serve to maximize the assignment of the rewarding function, this is not necessarily the best strategy for such a system. The reason behind this is the {\em greedy} behavior of such a strategy, (i.e. a behavior that always chooses the best rewarding next step). Such a behavior can achieve a local maximum regarding the rewarding function, but it doesn't search at all for possibly better solutions. The problem can be counteracted in many ways in a diagonalizing system, for example, assuming that {\em behaving randomly} is a parameter for such a system, and {\em sufficiency of resourses} is a noticeable condition, then diagonalization can relate them in the sense we have discussed in section 5 of \cite{partI}, causing a more randomized behavior under the condition of sufficiently gathered points of the rewarding function.

The remark implies that in order to develop computer systems that learn, in a lot of situations, evolving by means of diagonalization is expected to behave much better than programming.  The reason for this, lies onto the fact that learning depends upon a lot of rules that all the same have a lot of exceptions, as the one explained above. Notice for example, that one can program a self-editing system to diagonalize based on reward-punishment, yet as previously seen, this is not at all what one wants from it.}
\end{remark} 

\subsection{Negative diagonalization} Punishments bring into consideration the problem of whether it is possible for a diagonalizing system, to understand general causes behind them. The problem may be compared to the behavior of a self-editing system that has access not only to information about surviving codes, but also to information about non-surviving ones. 

Exactly as diagonalization may serve to deduce general behaviors for surviving from the mere information of what has survived up to now, {\em negative diagonalization} can be used to deduce general behaviors that should be avoided, judging from what has been avoided up to now. The procedure described as {\em punishment reaction} in paragraph \ref{punishmentReaction} may serve in this case to create a sample of already avoided recursors. Alternatively, such a sample may be considered as occurring by mere chance in a surviving sequence, shifting the problem of gathering the basic information of the appropriate sample, rather to a big population than to punishment reaction.

Before we present such a diagonal instruction, let us first introduce some terminology: By the term {\em accepting-rejecting} or simply {\em testing} algorithm, we mean an algorithm that answers only \textsf{True} or \textsf{False}. We are going to use the same terminology about codes that describe algorithms.

Thus, a diagonal instruction for generalizing rejection of recursors could read as follows:
\begin{equation} \label{negativeDiagonal} \textsf{\begin{tabular}{p{10cm}} Find a testing code $n,$  such that $\alg(n)(r) = \textsf{False},$ exactly for every rejected (or punished) code $r$ in the memory sequence.
\end{tabular}} \end{equation}

The meaning of \textsf{exactly} in the above diagonal instruction, is to ensure that such a rejecting code as $n,$ does not reject also useful codes. In other words, a negative diagonalization should aim to separate the codes to be rejected from any other code.

\subsubsection{An example of \eqref{negativeDiagonal}} \label{negativeExample} We aim to demonstrate here an example of the application of \eqref{negativeDiagonal}. 

Assume that we conduct a series of experiments, that are indicated by a noticeable for the system condition $E,$ (or by a condition that may become noticeable, in the sense of paragraph 5.3 of \cite{partI}), in which we ask  our system to output a specific kind of code, for example an integer. 

In each specific experiment of the series, punishing non-integer outputs and on the basis that a code $n$ of the accepting-rejecting algorithm 
\begin{equation} \label{negativeExampleEq} \alg(n)(r) = \begin{cases} \textsf{True}, & \text{ if $\alg(r)$ outputs an integer} \\ \textsf{False}, & \text{ if not} \end{cases}, \end{equation}
is simple enough, negative diagonalization \eqref{negativeDiagonal}, may ensure that in each such experiment, our system can adapt to non outputting other codes than integers. 

As a second step now, since in every case that $E$ holds, (that is in every experiment of the series), the system is forced to use \eqref{negativeExampleEq}, a code for
\[ \textsf{If $E$ holds, then use $n$ as a testing code,} \]
should fit the memory sequence, so that by diagonalization, the system can learn to use \eqref{negativeExampleEq} in each experiment of the series.

\subsubsection{Searching more efficiently} \label{efficientSearching} Let us begin by noticing the following:
\begin{remark} {\em Example \ref{negativeExample} may be adapted via addresses to other than the output module of the code, so that, under the necessary assumptions, such a system may understand the nature of the module in question, i.e. the nature of the codes that should be registered there.

Notice also that other than the nature of the code features (for example the range of a value) may be handled by the same way we used in \ref{negativeExample}} \end{remark}

Now, in many cases, if it happens to know the nature of a code, we can adapt ourselves to search among the codes of the same nature. For example, if, as above, this code is in fact an integer, then the code generator should be restricted to output only integers, thus searching more efficiently by adapting to the problem at hand. So, the following important question arises very naturally:
\begin{question} Is such a system able to adapt to negative diagonalization, by searching more efficiently? \end{question}

For the moment, we do not plan to investigate in detail the above question, yet one should notice that the module of the code generator is itself a subject of diagonalization due to the self-editing property. Combining this with {\em specialization} which we are going to introduce later on, we can have a hint towards this direction.

\subsubsection{Establishment of \eqref{negativeDiagonal}}It is interesting to notice here, that \eqref{negativeDiagonal} may be detected as a useful strategy, by simple diagonalization as it has been described in (4) of \cite{partI}, granted that our system has already established some simple codes of accepting-rejecting algorithms. 

The arguments here are similar to the ones presented in paragraphs 5.2 and 5.3 of \cite{partI}.

Indeed, if some simple codes of accepting-rejecting algorithms have already been established, and on the basis that these have to be correct, (since we are talking about surviving sequences), they have also to reject some of the recursors that have already been rejected, either by means of punishment reaction, or by mere chance. Thus \eqref{negativeDiagonal} should fit such a surviving sequence and therefore should be established by diagonalization, granted as usually that it has a simple enough code.

\begin{remark} {\em Establishment of the basic diagonal instruction (4) of \cite{partI} out of negative diagonalization is also possible, by using a testing code $n$ for which
\[ \alg(n)(r) = \begin{cases} \textsf{True}, & \text{if $r$ fits the memory sequence}\\ \textsf{False}, & \text{if not} \end{cases}. \]

The remark suggests that in order to materialize a diagonalizing system, it is probably better than programming diagonal instructions, to rely on an appropriate structure in combination with a great proliferating population that renders the possibility of creation of such instructions as significant. We will postpone the suggestions for such an appropriate structure for later on. 
}
\end{remark}


  \end{document}